\documentclass[runningheads]{llncs}
\usepackage{graphicx}
\usepackage{amsmath,amssymb} 
\usepackage{color}

\begin{document}
\title{3DContextNet: K-d Tree Guided Hierarchical Learning of Point Clouds Using Local and Global Contextual Cues} 

\titlerunning{3DContextNet}
%
\author{Wei Zeng, 
Theo Gevers}
%
\authorrunning{Wei Zeng and Theo Gevers}
%

\institute{Computer Vision Lab, University of Amsterdam\\
\email{w.zeng@uva.nl, th.gevers@uva.nl}}

\maketitle              
\begin{abstract}
Classification and segmentation of 3D point clouds are important tasks in computer vision. Because of the irregular nature of point clouds, most of the existing methods convert point clouds into regular 3D voxel grids before they are used as input for ConvNets. Unfortunately, voxel representations are highly insensitive to the geometrical nature of 3D data. More recent methods encode point clouds to higher dimensional features to cover the global 3D space. However, these models are not able to sufficiently capture the local structures of point clouds. 

Therefore, in this paper, we propose a method that exploits {\em both} local and global contextual cues imposed by the k-d tree. The method is designed to learn representation vectors progressively along the tree structure. Experiments on challenging benchmarks show that the proposed model provides discriminative point set features. For the task of 3D scene semantic segmentation, our method significantly outperforms the state-of-the-art on the Stanford Large-Scale 3D Indoor Spaces Dataset (S3DIS).

\end{abstract}
\section{Introduction}

Over the past few years, ConvNets have achieved excellent performance in different computer vision tasks such as image classification~\cite{LeNet5,imagenet1,imagenet2012}, object detection~\cite{rcnn,fast-rcnn,faster-rcnn} and semantic segmentation~\cite{rcnn,segnet,FCN,deconv}. 

3D imaging technology has also experienced a major progress. In parallel, a number of annotated large-scale 3D datasets have become publicly available, which are crucial for supervised 3D deep learning models. For example, ModelNet~\cite{3d-shapenets} and ShapeNet~\cite{shapenet} provide object-level man-made 3D models, whereas Stanford Large-Scale 3D Indoor Spaces Dataset~\cite{SIS3d} and ScanNet~\cite{scannet} are available as real 3D scene datasets. 

Most of the traditional work convert the irregular 3D data (point clouds) to regular formats like 2D projection images~\cite{MVCNN,deeppano,subvolumn} or 3D voxel grids~\cite{3d-shapenets,subvolumn,voxnet} as a pre-processing step.
Methods that employ 2D image projections of 3D models as their input, such as ~\cite{MVCNN,deeppano}, are well suited as inputs to 2D ConvNet architectures. However, the intrinsic 3D geometrical information is distorted by the 3D-to-2D projection. Hence, this type of methods are limited by the exploitation of 3D spatial connections between regions. While it might seem straightforward to extend 2D CNNs to process 3D data by utilizing 3D convolutional kernels, data sparsity and computational complexity are the restrictive factors of this type of approaches~\cite{3d-shapenets,voxnet,ORION,VRN}. 

\begin{figure}[t]
\includegraphics[scale=0.35]{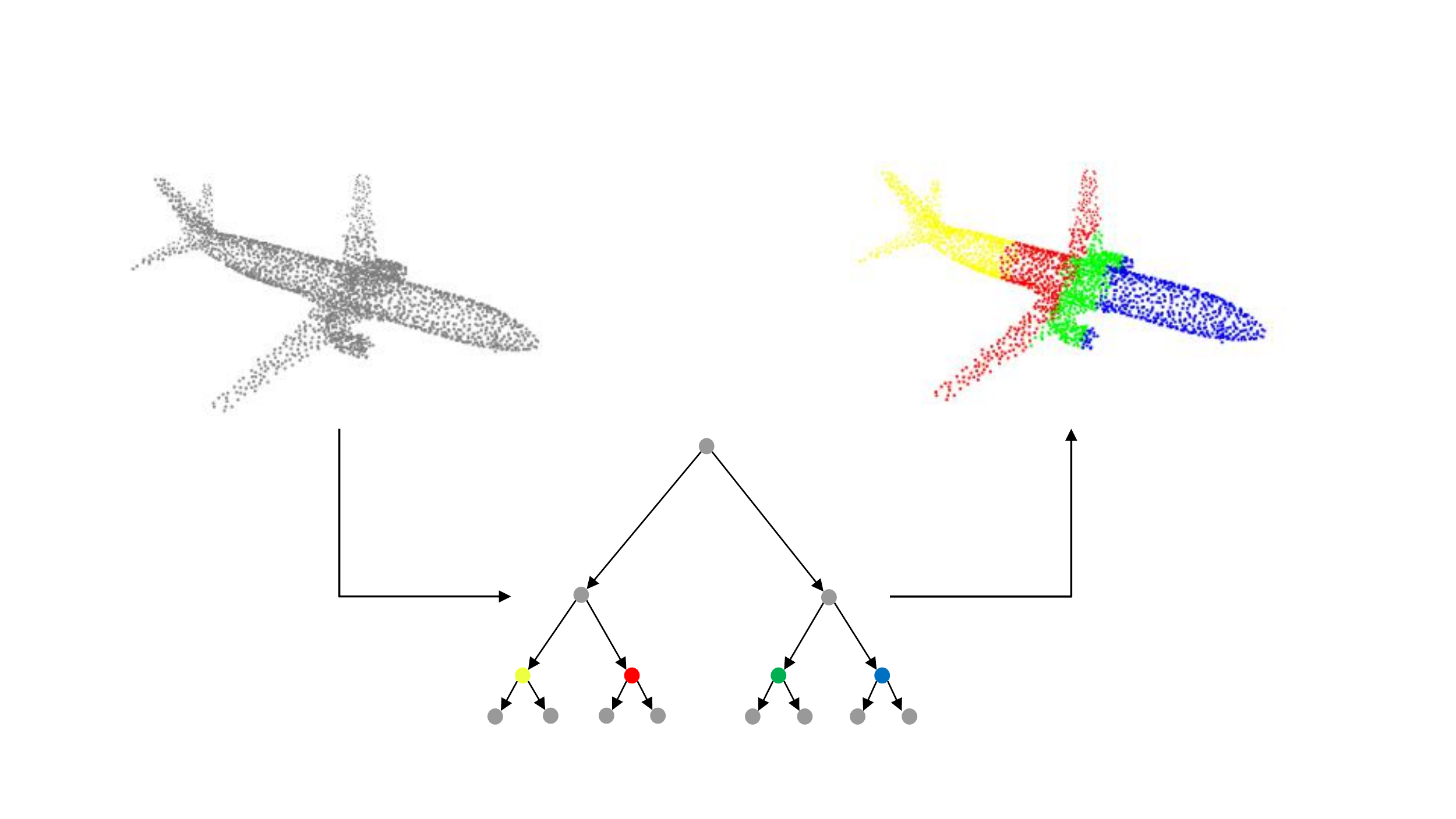}
\centering
\caption{Example of the implicit 3D space partition of a k-d tree. Colors of different local parts indicate different corresponding nodes in the k-d tree structure}
\label{fig:k-d tree structure}
\end{figure}

To fully exploit the 3D nature of point clouds, in this paper, the goal is to use the k-d tree structure~\cite{kd-tree} as the 3D data representation model, see Figure~\ref{fig:k-d tree structure}. Our method consists of two parts: {\em feature learning} and {\em aggregation}. The model exploits both local and global contextual information and aggregates point features to obtain discriminative 3D signatures in a hierarchical manner. In the feature learning stage, local patterns are identified by the use of an adaptive feature recalibration procedure, and global patterns are calculated as non-local responses of different regions at the same level. Then, in the feature aggregation stage, point features are merged hierarchically corresponding to the associated k-d tree structure in bottom-up fashion.

Our main contributions are as follows: (1) a novel 3D context-aware neural network is proposed for 3D point cloud feature learning by exploiting the implicit partition space of the k-d tree structure, (2) a novel method is presented to incorporate both local and global contextual information for point cloud feature learning, (3) for semantic segmentation, our method significantly outperforms the state-of-the-art on the challenging Stanford Large-Scale 3D Indoor Spaces Dataset(S3DIS)~\cite{SIS3d}.

\section{Related Work}

\begin{figure*}[t]
\includegraphics[scale=0.38]{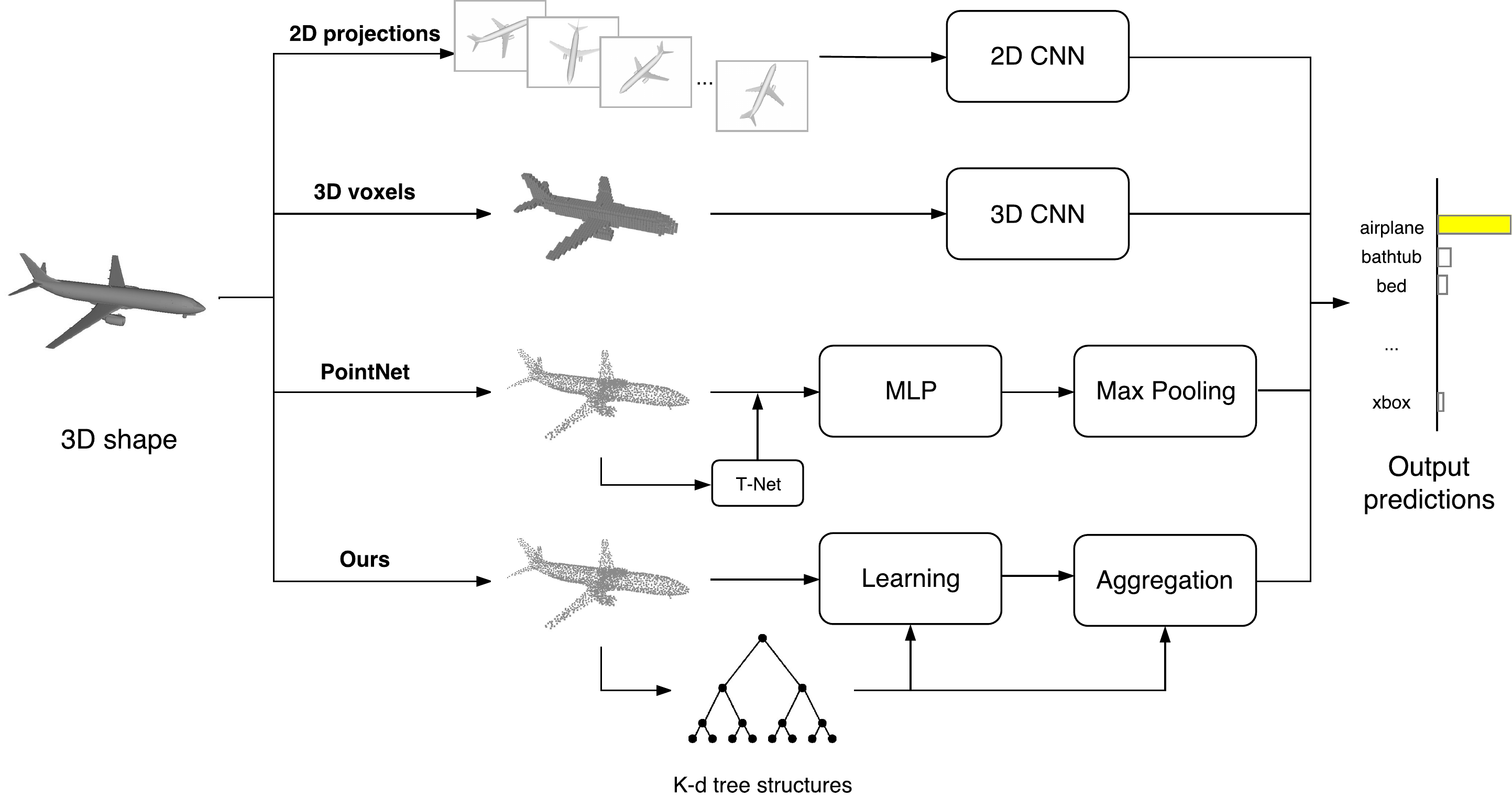}
\centering
\caption{Comparison to related work for the classification task. Our model is based on hierarchical feature learning and aggregation using the k-d tree structure}
\label{fig:comparison}
\end{figure*}

Previous work on ConvNets and volumetric models use different rasterization strategies. Wu et al. propose 3DShapeNets~\cite{3d-shapenets} using 3D binary voxel grids as input of a Convolutional Deep Belief Network. This is the first work to use deep ConvNets for 3D data processing. VoxNet~\cite{voxnet} proposes a 3D ConvNet architecture to integrate the 3D volumetric occupancy grid. ORION~\cite{ORION} exploits the 3D orientation to improve the results of voxel nets for 3D object recognition. Based on the ResNet~\cite{ResNet} architecture, Voxception-ResNet (VRN)~\cite{VRN} proposes a very deep architecture. OctNet~\cite{octnet} exploits the sparsity in the input data by using a set of unbalanced octrees where each leaf node stores a pooled feature representation. However, most of the volumetric models are limited by their resolution, data sparsity, and computational cost of 3D convolutions. 

Other methods rely on 2D projection images to represent the original 3D data and then apply 2D ConvNets to classify them. MVCNN~\cite{MVCNN} uses 2D rendered images of 3D shapes to learn representations of multiple views of a 3D model and then combines them to compute a compact descriptor. DeepPano~\cite{deeppano} converts each 3D shape to a panoramic view and uses 2D ConvNets to build classifiers directly from these panoramas. With well-designed ConvNets, this type of methods (2D projections from 3D) performs successfully in different shape classification and retrieval tasks. However, due to the 3D-to-2D projection, these methods are limited in exploring the full 3D nature of the data. In addition, ~\cite{s1,s2} exploits ConvNets to process non-Euclidean geometries. Moreover, Geodesic Convolutional Neural Networks (GCNN)~\cite{s2} apply linear and non-linear transformations to polar coordinates in a local geodesic system. However, these methods are limited to manifold meshes.

Only recently, a number of methods are proposed that apply deep learning directly to the raw 3D data (point clouds). PointNet~\cite{pointnet} is the pioneering work that directly processes 3D point sets in a deep learning setting. Nonetheless, since every point is treated equally, this approach fails in retaining the full 3D information. The modified version of PointNet, PointNet++~\cite{pointnet++}, abstracts local patterns by sampling representative points and recursively applies PointNet~\cite{pointnet} as a learning component to obtain the final representation. However, it directly discards the unselected points after each layer, and needs to sample points recursively at different scales which may yield relatively slow inference speed. Another recent work, Kd-Network~\cite{kd-network} uses a 3D indexing structure to perform the computation. The method employs parameter sharing and calculates representations from the leaf nodes to the roots. However, this method needs to sample the point clouds and to construct k-d trees for every iteration. Further, the method employs multiple k-d trees to represent a single object. It is split-direction-dependent and is negatively influenced by a change in rotation (3D object classification) and viewpoint (3D scene semantic segmentation).

In contrast to previous methods, our model is based on a hierarchical feature learning and aggregation pipeline. Our neural network structure exploits the local and global contextual cues which are inferred by the implicit space partition of the k-d tree. In this way, our model learn features, and calculates the representation vectors progressively using the associated k-d tree. Figure~\ref{fig:comparison} shows a comparison of related methods to our work for the classification task.



\section{Method}

\begin{figure*}[t]
\includegraphics[scale=0.29]{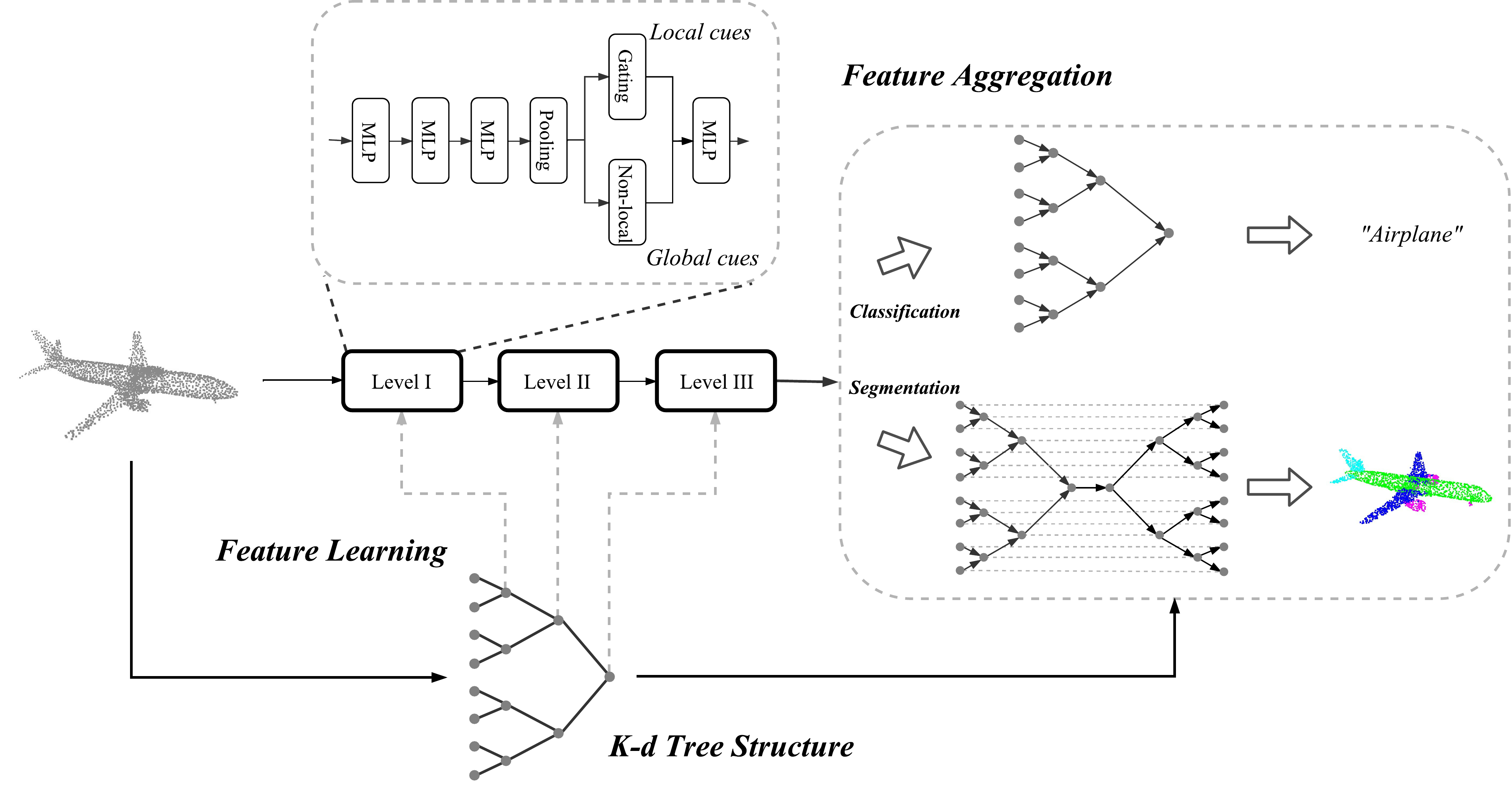}
\centering
\caption{3DContextNet architecture. 3D object point clouds are used to illustrate that our method is suitable for both 3D classification and segmentation tasks. The corresponding nodes of the k-d tree determine the receptive fields at different levels. For feature learning, both local and global contextual information is encoded for each level. The associated k-d tree forms the computational graph to compute the representation vectors progressively for feature aggregation}
\label{fig:our network architecture}
\end{figure*}

In this section, we describe our architecture, {\em 3DContextNet}, see Figure~\ref{fig:our network architecture}. First, the choice of the tree structure is motivated to subdivide the 3D space. Then, the feature learning stage is discussed that uses both local and global contextual cues to encode the point features. Finally, the feature aggregation stage is described that computes representation vectors progressively along the k-d trees.
\subsection{K-d Tree Structure: Implicit 3D Space Partition}
Our method is designed to capture both the local and global context by learning and aggregating point features progressively and hierarchically. Therefore, a representation model is required to partition 3D point clouds to encapsulate the latent relations between regions. To this end, the k-d tree structure~\cite{kd-tree} is chosen.

A k-d tree is a space partitioning structure which is constructed by recursively computing axis-aligned hyperplanes to divide point sets. In this paper, we choose the standard k-d tree construction to obtain balanced k-d trees from the 3D input point clouds/sets. The latent region subdivisions of the constructed k-d tree are used to capture the local and global contextual information of point sets. Each node, at a certain level, represents a local region at the same scale, whereas nodes at different levels represent subdivisions at corresponding scales. In contrast to the k-d network of~\cite{kd-network}, splitting directions and positions are not used for the tree construction. In this way, our method is more robust to jittering and rotation than~\cite{kd-network} which trains different affine transformations depending on the splitting directions of the nodes. 

The k-d tree structure can be used to search for k-nearest neighbors for each point to determine the local point adjacency and neighbor connectivity. Our approach uses the implicit local partitioning obtained by the k-d tree structure to determine the point adjacency and neighbor connectivity. 

In general, conventional ConvNets learn and merge nearby features at the same time enlarging the receptive fields of the network. Because of the non-overlapping partitioning of the k-d tree structure, in our method, learning and merging at the same time would decrease the size of the remaining points too fast. This may lead to a lack of fine geometrical cues which are factored out during the early merging stages. To this end, our approach divides the network architecture into two parts: {\em feature learning} and {\em aggregation}.

\subsection{Feature Learning Stage}
Given as input is a 3D point set with the corresponding k-d tree. The tree leaves contain the individual (raw) 3D points with their representation vectors, denoted by $X=\{x_{1},\dots, x_{n}\}\subseteq R^{F}$. For example, $F = 3$ denotes the initial vectors containing the 3D point coordinates. Features are directly learned from the raw point clouds without any pre-processing step. 
According to~\cite{Deepsets}, a function $S(X)$ is permutation invariant to the elements in $X$, if and only if it can be decomposed in the form of $\rho(\sum_{x\in X}\varphi(x))$, for a suitable transformation of $\rho$ and $\varphi$. We follow PointNet~\cite{pointnet}, where a point set is mapped to a discriminative vector as follows: 
\begin{equation}\label{eq1}
f(\{x_{1},\ldots,x_{n}\}) \approx g(h(x_{1}),\ldots,h(x_{n})),
\end{equation}
where $f: 2^{\mathbb{R}^N}\to\mathbb{R}$, $h: \mathbb{R}^N\to\mathbb{R}^K$ and $g: \underbrace{\mathbb{R}^K\times\ldots\times\mathbb{R}^K}_{n}\to\mathbb{R}$ is a symmetric function. 

In the feature learning stage, point features are computed at different levels hierarchically. For a certain level, we first process each point using shared multi-layer perceptron networks (MLP) as function $h$ in equation~\eqref{eq1}. Then, different local region representations are computed by a symmetric function, max pooling in our work, for the subdivision regions at the same level, as function $g$ in equation~\eqref{eq1}. Then, local and global contextual cues are calculated in parallel based on the local region representations. Note that both the local and global features are concatenated with the corresponding points to retain the number of points. 

\subsubsection{Local Contextual Cues: Adaptive Feature Recalibration }

To model the inter-dependencies between point features in the same region, we use the local region representations obtained from the symmetric function to perform adaptive feature recalibration~\cite{SENets}. All operations are adaptive to each local region, represented by a certain node in the k-d tree. The local region representation obtained by the symmetric function can be interpreted as a feature descriptor for the corresponding local region. A gating function is used with a sigmoid activation to capture the feature-wise dependencies. Point features in this local region are then rescaled by the activations to obtain the adaptive recalibrated output: 
\begin{equation}\label{eq2}
\tilde{y}_{i}=\sigma(g(Y))\cdot y_{i}, \qquad \quad i = 1, ..., m
\end{equation}
where $\sigma$ denotes the sigmoid activation and $g$ is the symmetric function to obtain the local region representation. $Y=\{y_{1}, \ldots, y_{m}\}$ is the point feature set of the local region and $m$ is the number of points in that region. In this way, feature dependencies are consolidated for each local region by enhancing informative features. As a result, we can obtaion more discriminative local patterns. Note that the activations act as feature weights and adaptively recalibrate point features for different local regions. 

\subsubsection{Global Contextual Cues: Non-local Responses}

Global contextual cues are based on the non-local responses to capture a greater range of dependencies. Intuitively, a non-local operation computes the response for one position as a weighted sum over the features for all positions in the input feature maps. A generic non-local operation~\cite{non-local} in deep neural networks is calculated by: 
\begin{equation}\label{eq3}
z_{i}=\frac{1}{C(x)}\sum_{\forall j}G(x_{i},x_{j})H(x_{j}),
\end{equation}
where $i$ is the index of the output position and $j$ is the index that enumerates all possible positions. In our case, $i$ represents a local region at a certain level and $j$ enumerates the number of local regions at the same level. Function $G$ denotes the relationships between $i$ and $j$. Further, function $H$ computes a representation of the input signal at position $j$. Then, the response is normalized by a factor $C(x)$. 

The k-d tree divides the input point set into different local regions. These are represented by different nodes of the tree. Larger range dependencies for different local regions at the same level are computed as non-local responses of the corresponding nodes of the tree. We consider $H$ as an MLP, and the pairwise function $G$ as an embedded Gaussian function: 
\begin{equation}\label{eq4}
G(x_{i},x_{j})=e^{\theta(x_{i})^{T}\phi(x_{j})},
\end{equation}
where $\theta(x_{i})$ and $\phi(x_{j})$ are two MLPs representing two embeddings. In this paper, the relationships between different nodes at the same level should be undirected, and hence $G(x_{i},x_{j})=G(x_{j},x_{i})$. Therefore, the two embeddings are the same i.e. $\theta = \phi$. The normalization factor is calculated by $C(x)=\sum_{\forall j}G(x_{i},x_{j})$. Note that this operation is different from a fully-connected layer. The non-local responses are based on the connections between different local regions, whereas fully-connected layers use learned weights. 

Due to our input format and architecture, the receptive fields of the convolutional kernels are always $1\times1$ in the feature learning stage. Following DenseNet~\cite{DenseNet}, to strengthen the information flow between layers, layers at the same level are connected (in the feature learning stage) with each other by concatenating all corresponding point features together. 
Such connections also lead to an implicit deep supervision which makes the network easier to train. The output of the feature learning stage has the same number of points as the input point set.

\subsection{Feature Aggregation Stage}

In the feature aggregation stage, the associated k-d tree structure is used to form the computational graph to progressively abstract over larger regions. For the classification task, the global signature is computed for the entire 3D model. For the semantic segmentation task, the outputs are the point labels. Instead of aggregating the information once over all points, the more discriminative features are computed in a bottom-up manner. The representation vector of a non-leaf node at a certain level is computed from its children nodes by MLPs and the symmetric function. To that end, max pooling is used as the symmetric function. 

For classification, by using this bottom-up and hierarchical approach, more discriminative global signatures are obtained. This procedure corresponds to a ConvNet in which the representation of a certain location is computed from the representations of nearby locations at the previous layers by a series of convolutions and pooling operations. Our architecture is able to progressively capture features at increasingly larger scales. Features at lower levels have smaller receptive fields, whereas features at higher levels have larger receptive fields. That is due to the data-dependent partition of the k-d tree structure. Additionally, our model is invariant to the input order of the point sets, because the aggregating direction is along the k-d tree structure, which is invariant to input permutations. 

For the semantic segmentation task, the k-d tree structure is used to represent an encoder-decoder architecture with skip connections to link the related layers. The input of the feature aggregation stage is the point feature set in which the representation of each point encapsulates both local and global contextual information at different scales. The output is a semantic label for each point. 

In conclusion, our architecture fully utilizes the local and global contextual cues in the feature learning stage. It calculates the representation vectors hierarchically in the feature aggregation stage. Hence, with k-d tree guided hierarchical learning, our 3DContextNet can obtain discriminative features for point clouds. 

\subsection{Discussion}
Our method is related to PointNet~\cite{pointnet} which encodes the coordinates of each point to higher dimensional features. However, by its design, this method is not able to sufficiently capture the local patterns in 3D space. More recently, PointNet++~\cite{pointnet++} is proposed which abstracts local patterns by selecting representative points in a metric space and recursively applies PointNet as a local feature learner to obtain features of the whole point set. In fact, the method handles the non-uniform point sampling problem. However, the set of abstraction layers need to sample the point sets multiple times at different scales which leads to a relative slow inference speed. Further, only the selected points are preserved. Others are directly discarded after each layer which causes the loss of fine geometric details. Another recent work, K-d network~\cite{kd-network} performs linear and non-linear transformations and share the transformation parameters corresponding to the splitting directions of each node in the k-d tree. The input of this method is the constructed k-d trees. It needs to calculate the representation vectors for all the nodes of the associated tree structure. For each node at a certain level, the input is the representation vectors of the two previous nodes. The method heavily depends on the splitting direction of each node to train different multiplicative transformations at each level. Hence, the method is not invariant to rotation. Furthermore, point cloud sampling and k-d tree fitting during every iteration lead to slow training and inference speed.

\subsection{Implementation Details}

Our 3DContextNet model deals with point clouds of a fixed size $N = 2^D$ where $D$ is the depth of the corresponding balanced k-d tree. Point clouds of different sizes can be converted to the same size using sub- or oversampling. In our experiments, not all the levels of the k-d tree are used. For simplicity and efficiency reasons, this number is $L=3$ for both the feature learning and aggregation stage. The receptive fields (number of points) for each level in the feature learning stage are 32 - 64 - 128 for the classification tasks and 32 - 128 - 512 for the segmentation tasks.

In the feature learning stage, the sizes of the shared MLPs are (64, 64, 128, 128) - (64, 64, 256, 256) - (64, 64, 512, 512) for the three levels, respectively. The size of MLPs for $\theta$ and $H$ are 64 - 128 - 256 and 128 - 256 - 512, respectively. Dense connections are applied within each level before the max-pooling layer. In the feature aggregation stage, the MLPs and pooling operations are used recursively to progressively abstract the discriminative representations. For the classification task, the sizes of the MLPs are (1024) - (512) - (256), respectively. For the segmentation task, like the hourglass shape, the sizes of the MLPs are (1024) - (512) - (256) - (256) - (512) - (1024), respectively. The output is then processed by two fully-connected layers with size 256. Dropout is applied after each fully-connected layer with a ratio of $0.5$.  


\section{Experiments}

In this section, we evaluate our 3DContextNet on different 3D point cloud datasets. First, it is shown that our model significantly outperforms state-of-the-art methods for the task of semantic segmentation on the Stanford Large-Scale 3D Indoor Spaces Dataset~\cite{SIS3d}. Then, it is shown that our model provides competitive results for the task of 3D object classification on the ModelNet40 dataset~\cite{3d-shapenets} and the task of 3D object part segmentation on the ShapeNet part dataset~\cite{shapenet}.

\subsection{3D Semantic Segmentation of Scenes}

\begin{table}[t]
\begin{center}
\caption{3D semantic segmentation results on the Stanford Large-Scale 3D Indoor Spaces Dataset (S3DIS). Our method outperforms previous state-of-the-art methods by a large margin}
\label{table:overall}
\begin{tabular}{l | c | c | c}
\hline
      & mean IoU & overall accuracy & avg. class accuracy  \\
\hline
Baseline~\cite{pointnet} & 20.1 & 53.2 & - \\

PointNet~\cite{pointnet} & 47.6 & 78.5 &  66.2\\

MS + CU(2)~\cite{iccvworkshop} & 47.8 & 79.2 & 59.7\\

G + RCU~\cite{iccvworkshop} & 49.7 & 81.1 & 66.4 \\

PointNet++~\cite{pointnet++} & 53.2 & 83.0 & 70.5 \\
\hline
Ours &  \textbf{55.6}  &  \textbf{84.9}  &  \textbf{74.5} \\
\hline
\end{tabular}
\end{center}
\end{table}

Our network is evaluated on the \emph{Stanford Large-Scale 3D Indoor Spaces} (S3DIS) dataset~\cite{SIS3d,3djoint} for 3D semantic segmentation task. The dataset contains 6 large scale indoor areas and each point is labeled with one of the 13 semantic categories, including 5 types of furniture (\emph{board}, \emph{bookcase}, \emph{chair}, \emph{sofa} and \emph{table}) and 7 building elements (\emph{ceiling}, \emph{beam}, \emph{door}, \emph{wall}, \emph{window}, \emph{column} and \emph{floor}) plus \emph{clutter}. We follow the same setting as in~\cite{pointnet} and use a 6-fold cross validation over all the areas.

Our method is compared with the baseline by PointNet~\cite{pointnet} and the recently introduced MS+CU and G+RCU models~\cite{iccvworkshop}. We also produce the results of PointNet++~\cite{pointnet++} for this dataset. During training, we use the same pre-processing as in~\cite{pointnet}. We first split rooms into blocks of $1m\times1m$ and represent each point by a 9-dimensional vector containing coordinates ($x$, $y$, $z$), the color information $RGB$ and the normalized position ($x'$, $y'$, $z'$).  The baseline extracts the same 9-dim local features and three additional ones: local point density, local curvature and normals. The standard MLP is used as the classifier. PointNet~\cite{pointnet} computes the global point cloud signature and feeds it back to per point features. In this way, each point representation incorporates both local and global information. Recent work by~\cite{iccvworkshop} proposes two models that enlarge the receptive field over the 3D scene. The motivation is to incorporate both the input-level context and the output-level context. MS+CU represents the multi-scale input block with a consolidation unit model, while G+RCU stands for the grid-blocks in combination with a recurrent consolidation block model. PointNet++~\cite{pointnet++} exploits metric space distances to build a hierarchical grouping of points and abstracts the features progressively. Results are shown in Table \ref{table:overall}. 
A significance test is conducted between our results and the state-of-the-art results obtained by PointNet++~\cite{pointnet++}. The p-value equals to 0.0122 in favor of our method.

We also compare the mean IoU for each semantic class with $XYZ-RGB$ and only with $XYZ$ as input, see Table \ref{table:IoU per semantic class} and Table \ref{table:IoU per semantic class without color} respectively. We obtain state-of-the-art results in mean IoU and for most of the individual classes for both $XYZ-RGB$ and $XYZ$ input. 
The reason of obtaining comparable results with PointNet++~\cite{pointnet++} for furnitures is that the k-d tree structure is computed along the axes. Therefore, it may be inefficient for precise prediction near the splitting boundaries, especially for relatively small objects. Note that our model using only geometry information (i.e. $XYZ$) achieves better results than the original PointNet method using both geometry and color/appearance information.

\begin{table*}[t]
\begin{center}
\caption{IoU per semantic class for the S3DIS dataset with $XYZ-RGB$ as input. It can be derived that our method obtains the state-of-the-art results in mean IoU and for most of the individual classes}
\label{table:IoU per semantic class}
\scalebox{0.7}{
\begin{tabular}{l|c|c|c|c|c|c|c|c|c|c|c|c|c|c}
\hline
\    & mean IoU & Ceiling & Floor & Wall  & Beam & Column & Window & Door & Table & Chair & Sofa & Bookcase & Board & clutter \\
\hline
PointNet~\cite{pointnet} & 47.6 & 88.0 & 88.7 & 69.3 & 42.4 & 23.1 & 47.5 & 51.6 & 54.1 & 42.0 & 9.6 & 38.2 & 29.4 & 35.2 \\

MS + CU(2)~\cite{iccvworkshop} & 47.8 & 88.6 & \textbf{95.8} & 67.3 & 36.9 & 24.9 & 48.6 & 52.3 & 51.9 & 45.1 & 10.6 & 36.8 & 24.7 & 37.5\\

G + RCU~\cite{iccvworkshop} & 49.7 & 90.3  & 92.1  &  67.9  & 44.7 & 24.2 & 52.3 & 51.2 & 58.1 & 47.4 & 6.9 & 39.0 & 30.0 & 41.9 \\

PointNet++~\cite{pointnet++} & 53.2 & 90.2 & 91.7 & 73.1 & 42.7 & 21.2 & 49.7 & 42.3 & 62.7 & \textbf{59.0} & 19.6 & \textbf{45.8} & \textbf{48.2} & 45.6 \\
\hline
Ours & \textbf{55.6} & \textbf{92.6} & 93.1 & \textbf{73.9} & \textbf{52.9} & \textbf{35.0} & \textbf{55.8} & \textbf{57.5} & \textbf{62.9} & 49.0 & \textbf{22.0} & 42.8 & 39.8 & \textbf{45.8} \\
\hline
\end{tabular}
}
\end{center}
\end{table*}

\begin{table*}[t]
\begin{center}
\caption{IoU per semantic class for the S3DIS dataset using only $XYZ$ input features (no color/appearance). It is shown that our method provides comparable results in mean IoU and for all individual classes even without color/appearance information}
\label{table:IoU per semantic class without color}
\scalebox{0.72}{
\begin{tabular}{l|c|c|c|c|c|c|c|c|c|c|c|c|c|c}
\hline
\    & mean IoU & Ceiling & Floor & Wall  & Beam & Column & Window & Door & Table & Chair & Sofa & Bookcase & Board & clutter \\
\hline
PointNet~\cite{pointnet} & 40.0 & 84.0 & 87.2 & 57.9 & 37.0 & 19.6 & 29.3 & 35.3 & 51.6 & 42.4 & 11.6 & 26.4 & 12.5 & 25.5 \\

MS + CU(2)~\cite{iccvworkshop} & 43.0 & 86.5 & \textbf{94.9} & 58.8 & 37.7 & 25.6 & 28.8 & 36.7 & 47.2 & 46.1 & 18.7 & 30.0 & 16.8 & 31.2 \\

PointNet++~\cite{pointnet++} & 47.0 & 88.0 & 92.4 & \textbf{64.7} & 37.7 & 16.8 & 31.0 & \textbf{41.1} & \textbf{59.6} & \textbf{52.0} & \textbf{29.4} & \textbf{42.2} & \textbf{19.2} & 36.9 \\

\hline
Ours & \textbf{48.6} & \textbf{90.5} & 92.8 & 63.6 & \textbf{49.4} & \textbf{31.2} & \textbf{44.2} & 37.8 & \textbf{59.6} & 50.6 & 17.7 & 38.7 & 17.3 & \textbf{37.9}   \\
\hline
\end{tabular}
}
\end{center}
\end{table*}

\begin{figure*}[t]
\includegraphics[scale=0.44]{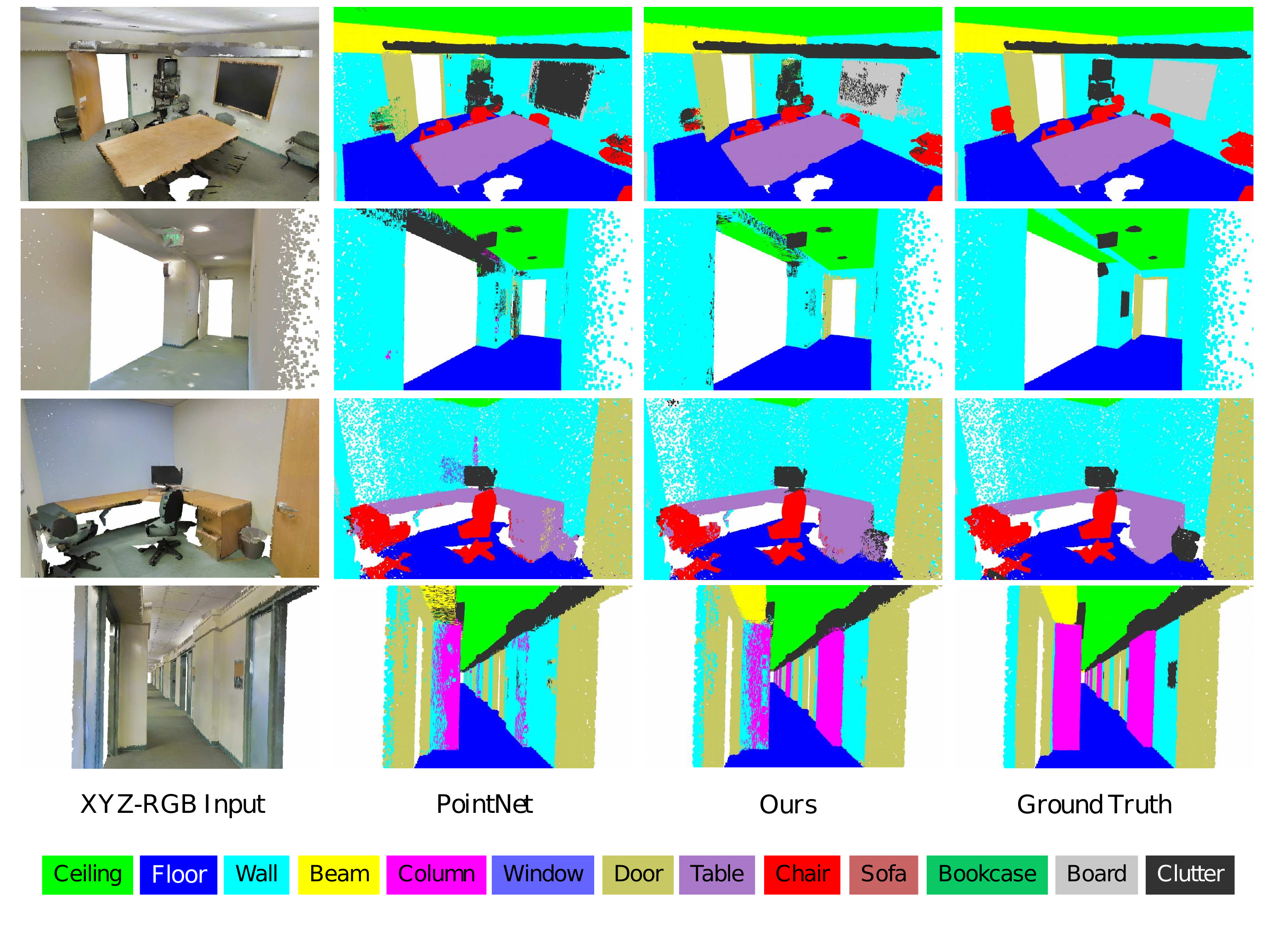}
\centering
\caption{Qualitative results for 3D indoor semantic segmentation. Results for the S3DIS dataset with $XYZ-RGB$ as input. From left to right: the input point cloud, the results of PointNet, our results, and the ground truth semantic labels. Our model obtains more consistent and less noisy predictions}
\label{fig:Qualitative results for semantic segmentation}
\end{figure*}

A number of qualitative results are presented in Figure~\ref{fig:Qualitative results for semantic segmentation} for the 3D indoor semantic segmentation task. It can be derived that our method provides more precise predictions for local structures. It shows that our model exploits both local and global contextual cues to learn discriminative features to achieve proper semantic segmentation. Moreover, our model size is less than 160 MB and average inference time is less than 70 ms per block, which makes our method suitable for large scale point cloud analysis.

\subsubsection{Ablation Study}

\begin{table}[t]
\caption{Effectiveness of different components of our architecture. We use the sixth fold setting of [24] for S3DIS as our training/testing split}
\label{table:ablation}
\begin{center}
\scalebox{0.9}{
\begin{tabular}{l | c | c }
\hline
      & mean IoU & overall accuracy \\
\hline

PointNet & 63.9 & 86.0 \\

PointNet++ & 69.1 & 90.1 \\

\hline
\hline

Baseline with k-d tree guided & 68.1 & 89.2 \\

Only Progressively Aggregation & 68.3 & 88.9 \\

Only Global Cues & 68.7 & 88.9 \\

Only Local Cues & 69.9 & 89.8 \\

\hline

Global Cues and Progressively Aggregation & 69.8 & 89.9 \\

Local Cues and Progressively Aggregation & 71.2 & 90.4 \\

Local and Global Cues & 71.5 & 90.1 \\

\hline

All & \textbf{72.0} & \textbf{90.6} \\

\hline
\end{tabular}
}
\end{center}
\end{table}

In this section, experiments are conducted to validate the effects of the different components of our proposed architecture for 3D semantic segmentation task. The baseline is the model corresponding to the vanilla PointNet, but utilizing the k-d tree partitioning to guide the feature learning stage. For a certain level, max-pooling is used to obtain different local region representations which are concatenated with the corresponding point features. We also trained models with different sets of components to test the effectiveness of our approach. We use the sixth fold setting of [24] for S3DIS as our experiment setting (i.e. we test on Area 6 and train on the rest). Results are reported in Table \ref{table:ablation}. Experimental results show that: (1) with k-d tree guided hierarchical feature learning, the baseline obtains better results than PointNet. Hence, local structures do help, (2) local contextual cues boost the performance the most, indicating that local neighborhoods of points contain fine-grained structure information, (3) any single combination of two components increases the performance and combining all of them provides state-of-the-art 3D semantic segmentation results.

\subsection{3D Object Classification and Part Segmentation}

We evaluate our method on the ModelNet40 shape classification benchmark~\cite{3d-shapenets}. The dataset contains a collection of 3D CAD models of 40 categories. We use the official split consisting of 9843 examples for training and 2468 for testing. Using the same experimental settings of~\cite{pointnet}, we convert the CAD models to point sets by uniformly sampling (1024 points in our case) over the mesh faces. Then, these points are normalized to have zero mean and unit sphere. We also randomly rotate the point sets along the $z$-axis and jitter the coordinates of each point by Gaussian noise for data augmentation during training.

It can be derived from Table \ref{table: 3D object classification}, that our model outperforms PointNet~\cite{pointnet}. 
Our model has competitive performance compared to PointNet++. However, our method is much faster in inference time. Table \ref{table: Comparison of the model sizes and the inference time} summarizes the comparison of time and space computations between PointNet, PointNet++ and our proposed method. We measure forward pass time with a batch size of 8 using TensorFlow 1.1. PointNet has the best time efficiency, but our model is faster than PointNet++ while keeping a comparable classification performance.

\begin{table}[t]
\begin{center}
\caption{3D object classification results on ModelNet40. The result of our model outperforms PointNet and is comparable to PointNet++}
\label{table: 3D object classification}
\scalebox{0.9}{
\begin{tabular}{l|c|c}
\hline
Method & Input & Accuracy (\%) \\
\hline
DeepPano~\cite{deeppano} & image & 77.6\\
MVCNN~\cite{MVCNN} & image & 90.1\\
MVCNN-MultiRes~\cite{subvolumn} & image & 91.4\\
3DShapeNets~\cite{3d-shapenets} & voxel & 77\\
VoxNet~\cite{voxnet} & voxel & 83\\
Subvolume~\cite{subvolumn} & voxel & 89.2\\
PointNet (vanilla)~\cite{pointnet} & point cloud & 87.2\\
PointNet~\cite{pointnet} & point cloud & 89.2\\
K-d network~\cite{kd-network} & point cloud & 90.6\\
PointNet++~\cite{pointnet++} & point cloud & 90.7\\
PointNet++ (with normal)~\cite{pointnet++} & point cloud & 91.9\\
\hline
Ours & point cloud & 90.2\\
Ours (with normal) & point cloud &  91.1 \\
\hline
\end{tabular}
}
\end{center}
\end{table}

\begin{table*}[t]
\begin{center}
\caption{Comparison of the model sizes and the inference time for the classification task. Our model is faster than PointNet++ while keeping comparable classification performance}
\label{table: Comparison of the model sizes and the inference time}
\scalebox{0.7}{
\begin{tabular}{l|c|c|c|c|c}
\hline
  &  PointNet~\cite{pointnet}  & PointNet++(SSG)~\cite{pointnet++} & PointNet++(MSG)~\cite{pointnet++} & PointNet++(MRG)~\cite{pointnet++} & 3DContextNet \\
\hline
Model size (MB) & 40 &  8.7 & 12 & 24  & 56.8\\

Forward time (ms) & 25.3  &  82.4 & 163.2 & 87.0& 45.9 \\
\hline
\end{tabular}
}
\end{center}
\end{table*}

We also evaluate our method on the ShapeNet part dataset~\cite{shapenet}. The dataset contains 16881 CAD models of 16 categories. Each category is annotated with 2 to 6 parts. There are 50 different parts annotated in total. We use the official split for training and testing. In this dataset, both the number of shapes and the parts within categories are highly imbalanced. Therefore, many previous methods train their network on every category separately. Our network is trained across categories.

\begin{figure*}[t]
\includegraphics[scale=0.3]{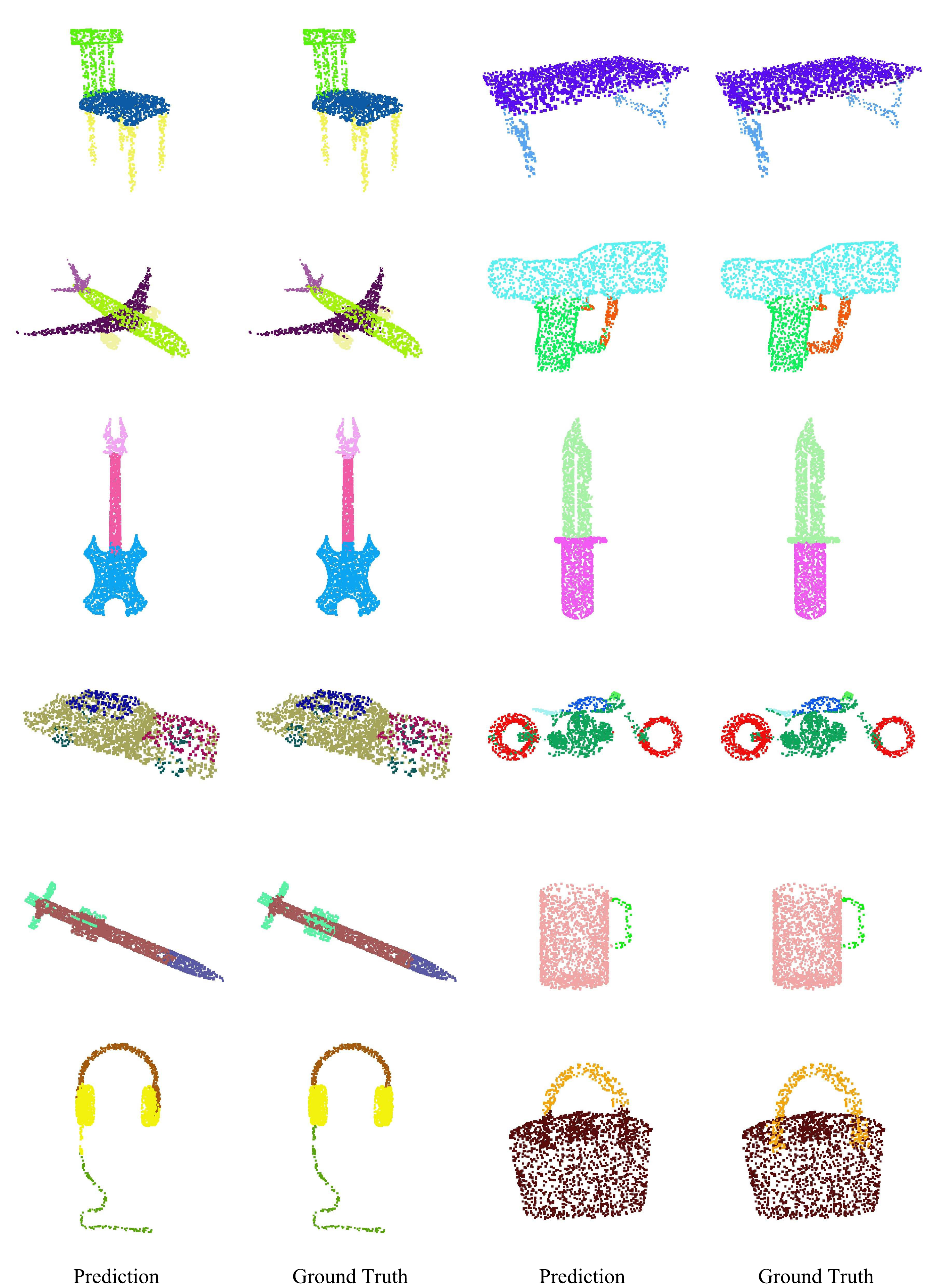}
\centering
\caption{Qualitative results for the 3D object part segmentation task. For each group from left to right: the prediction and the ground truth}
\label{fig:Qualitative results for 3D object part segmentation}
\end{figure*}

We compare our model with two traditional learning based techniques Wu~\cite{wu} and Yi~\cite{yi}, the volumetric deep learning baseline (3DCNN) in PointNet~\cite{pointnet}, as well as state-of-the-art approaches of SSCNN~\cite{sscnn} and PointNet++~\cite{pointnet++}, see Table \ref{table: 3D object part segmentation}. The point intersection over union for each category as well as the mean IoU are reported. In comparison to PointNet, our approach performs better on most of the categories, which proves the importance of local and global contextual information. See Figure~\ref{fig:Qualitative results for 3D object part segmentation} for a number of qualitative results for the 3D object part segmentation task.

\begin{table*}
\begin{center}
\caption{3D object part segmentation results on ShapeNet part dataset}
\label{table: 3D object part segmentation}
\scalebox{0.65}{
\begin{tabular}{l|c|c|c|c|c|c|c|c|c|c|c|c|c|c|c|c|c}
\hline
              & mean & airplane & bag & cap & car  & chair & earphone & guitar & knife & lamp & laptop & motor & mug & pistol & rocket & skateboard & table \\
\hline
\#shapes       &   & 2690 & 76 & 55 & 898 & 3758 & 69 & 787 & 392 & 1547 & 451 & 202 & 184 & 283 & 66 & 152 & 5271  \\
\hline
Wu~\cite{wu}             & - & 63.2 & -  & -  & -   & 73.5 &  - & -   & -   & 74.4 & -   & -   & -   & -   & -  &  -  & 74.8  \\
K-d Networks~\cite{kd-network}& 77.2 &79.9  &71.2&80.9&68.8 &88.0  &72.4&88.9 &86.4 &79.8  &94.9 &55.8 &86.5 &79.3 &50.4&71.1 &80.2   \\
3DCNN~\cite{pointnet}       & 79.4 &75.1  &72.8&73.3&70.0 &87.2  &63.5&88.4 & 79.6&74.4  &93.9 &58.7 &91.8 &76.4 &51.2&65.3 &77.1   \\ 
Yi~\cite{yi}          & 81.4 & 81.0 &78.4&77.7&75.7 &87.6  &61.9&92.0 & 85.4&82.5  &95.7 & 70.6&91.9 & 85.9&53.1&69.8 &75.3   \\
PointNet~\cite{pointnet}    & 83.7 &83.4  &78.7&82.5&74.9 &89.6  &73.0&91.5 &85.9 &80.8  &95.3 &65.2 &93.0 &81.2 &57.9&72.8 &80.6   \\
SSCNN~\cite{sscnn}       & 84.7 &81.6  &81.7&81.9&75.2 &90.2  &74.9&93.0 &86.1 &84.7  &95.6 &66.7 &92.7 &81.6 &60.6&82.9 &82.1   \\
PointNet++~\cite{pointnet++} & 85.1 & 82.4 & 79.0 & 87.7 & 77.3 & 90.8 & 71.8 & 91.0 & 85.9 & 83.7 & 95.3 & 71.6 & 94.1 & 81.3 & 58.7 & 76.4 & 82.6 \\
\hline
Ours        & 84.3 & 83.3 & 78.0  & 84.2  & 77.2  & 90.1  & 73.1 & 91.6 & 85.9  & 81.4 & 95.4   & 69.1 & 92.3 & 81.7  & 60.8 & 71.8 & 81.4 \\
\hline
\end{tabular}
}
\end{center}
\end{table*}

\section{Conclusion}

In this paper, we proposed a deep learning architecture that exploits the local and global contextual cues imposed by the implicit space partition of the k-d tree for feature learning, and calculate the representation vectors progressively along the associated k-d tree for feature aggregation. Large scale experiments showed that our model outperformed existing state-of-the-art methods for semantic segmentation task.  Further, the model obtained comparable results for 3D object classification and 3D part segmentation. 

In the future, other hierarchical 3D space partition structures can be studied as the underlying structure for the deep net computation and the non-uniform point sampling issue needs to be taken into consideration.


%
%
%
\bibliographystyle{splncs04}
\bibliography{egbib}
\end{document}